\RequirePackage{snapshot}

\documentclass{article}

\usepackage{microtype}
\usepackage{graphicx}
\usepackage{booktabs}

\usepackage{hyperref}

\usepackage{amsmath}
\usepackage{subfiles}
\usepackage{amssymb}
\usepackage{subcaption}
\usepackage{mycustom}

\usepackage[accepted]{icml2020}

\usepackage{cleveref}

\icmltitlerunning{The Conditional Entropy Bottleneck}

\title{The Conditional Entropy Bottleneck}

\author{%
Ian Fischer \\
Google Research \\
\texttt{iansf@google.com} \\
}

\begin{document}

\twocolumn[
\maketitle
\icmlkeywords{Information Bottleneck, Robustness, Adversarial Robustness, CEB, Machine Learning}
\vskip -0.1in
]
\begin{abstract}
Much of the field of Machine Learning exhibits a prominent set of failure modes, including vulnerability to adversarial examples, poor out-of-distribution (OoD) detection, miscalibration, and willingness to memorize random labelings of datasets.
We characterize these as failures of \textit{robust generalization}, which extends the traditional measure of generalization as accuracy or related metrics on a held-out set.
We hypothesize that these failures to robustly generalize are due to the learning systems retaining \textit{too much} information about the training data.
To test this hypothesis, we propose the \textit{Minimum Necessary Information} (MNI) criterion for evaluating the quality of a model.
In order to train models that perform well with respect to the MNI criterion, we present a new objective function, the \textit{Conditional Entropy Bottleneck} (CEB), which is closely related to the \textit{Information Bottleneck} (IB).
We experimentally test our hypothesis by comparing the performance of CEB models with deterministic models and Variational Information Bottleneck (VIB) models on a variety of different datasets and robustness challenges.
We find strong empirical evidence supporting our hypothesis that MNI models improve on these problems of robust generalization.
\end{abstract}

\section{Introduction}

Despite excellent progress in classical generalization (e.g., accuracy on a held-out set), the field of Machine Learning continues to struggle with the following issues:
\begin{itemize}[leftmargin=1em]
  \item \textbf{Vulnerability to adversarial examples.}
  Most machine-learned systems are vulnerable to adversarial examples.
  Many defenses have been proposed, but few have demonstrated robustness against a powerful, general-purpose adversary.
  Many proposed defenses are ad-hoc and fail in the presence of a concerted attacker~\citep{carlini2017adversarial,athalye2018obfuscated}.
  \item \textbf{Poor out-of-distribution detection.}
  Most models do a poor job of signaling that they have received data that is substantially different from the data they were trained on.
  Even generative models can report that an entirely different dataset has higher likelihood than the dataset they were trained on~\citep{oodlikelihood}.
  Ideally, a trained model would give less confident predictions for data that was far from the training distribution (as well as for adversarial examples).
  Barring that, there would be a clear, principled statistic that could be extracted from the model to tell whether the model \textit{should} have made a low-confidence prediction.
  Many different approaches to providing such a statistic have been proposed~\citep{oncalibration,ensemble,baseline,odin,confidence,learningconfidence}, but most seem to do poorly on what humans intuitively view as obviously different data.
  \item \textbf{Miscalibrated predictions.}
  Related to the issues above, classifiers tend to be overconfident in their predictions~\citep{oncalibration}.
  Miscalibration reduces confidence that a model's output is fair and trustworthy.
  \item \textbf{Overfitting to the training data.}
  \citet{rethinkinggeneralization} demonstrated that classifiers can memorize fixed random labelings of training data, which means that it is possible to learn a classifier with perfect \textit{inability} to generalize.
  This critical observation makes it clear that a fundamental test of generalization is that the model should \textit{fail} to learn when given what we call \textit{information-free} datasets.
\end{itemize}

We consider these to be problems of \textit{robust generalization}, which we define and discuss in~\Cref{sec:robust}.
In this work, we hypothesize that these problems of robust generalization all have a common cause: models retain \textit{too much} information about the training data.
We formalize this by introducing the \textit{Minimum Necessary Information} (MNI) criterion for evaluating a learned representation (\Cref{sec:mni}).
We then introduce an objective function that directly optimizes the MNI, the \textit{Conditional Entropy Bottleneck} (CEB) (\Cref{sec:ceb}) and compare it with the closely-related \textit{Information Bottleneck} (IB) objective \citep{tishby2000information} in~\Cref{sec:vib}.
In~\Cref{sec:models}, we describe practical ways to optimize CEB in a variety of settings.

Finally, we give empirical evidence for the following claims:
\begin{itemize}[leftmargin=1em]
  \item \textbf{Better classification accuracy.}
  MNI models can achieve superior accuracy on classification tasks than models that capture either more or less information than the minimum necessary information~(\Cref{sec:classification,sec:cifar}).
  \item \textbf{Improved robustness to adversarial examples.}
  Retaining excessive information about the training data results in vulnerability to a variety of whitebox and transfer adversarial examples.
  MNI models are substantially more robust to these attacks~(\Cref{sec:adversarial,sec:cifar}).
  \item \textbf{Strong out-of-distribution detection.}
  The CEB objective provides a useful metric for out-of-distribution (OoD) detection, and CEB models can detect OoD examples as well or better than non-MNI models~(\Cref{sec:ood}).
  \item \textbf{Better calibration.}
  MNI models are better calibrated than non-MNI models~(\Cref{sec:calibration}).
  \item \textbf{No memorization of information-free datasets.}
  MNI models fail to learn in information-free settings, which we view as a minimum bar for demonstrating robust generalization~(\Cref{sec:generalization}).
\end{itemize}

\section{Robust Generalization}
\label{sec:robust}

In classical generalization, we are interested in a model's performance on held-out data on some task of interest, such as classification accuracy.
In \textit{robust generalization}, we want:
\textbf{(RG1)} \emph{to maintain the model's performance in the classical generalization setting};
\textbf{(RG2)} \emph{to ensure the model's performance in the presence of an adversary (unknown at training time)};
and \textbf{(RG3)} \emph{to detect adversarial and non-adversarial data that strongly differ from the training distribution}.

Adversarial training approaches considered in the literature so far~\citep{fgm,kurakin2016adversarialml,madry2017towards} violate \textbf{(RG1)}, as they typically result in substantial decreases in accuracy.
Similarly, provable robustness approaches (e.g., \citet{cohen2019certified,wong2018scaling}) provide guarantees for a particular adversary known at training time, also at a cost to test accuracy.
To our knowledge, neither approaches provide any mechanism to satisfy \textbf{(RG3)}.
On the other hand, approaches for detecting adversarial and non-adversarial out-of-distribution (OoD) examples~\citep{oncalibration,ensemble,baseline,odin,confidence,learningconfidence} are either known to be vulnerable to adversarial attack~\citep{carlini2017adversarial,athalye2018obfuscated},
or do not demonstrate that the approach provides robustness against unknown adveraries, both of which violate \textbf{(RG2)}.

Training on information-free datasets~\citep{rethinkinggeneralization} provides an additional way to check if a learning system is compatible with \textbf{(RG1)}, as memorization of such datasets necessarily results in maximally poor performance on any test set.
Model calibration is not obviously a necessary condition for robust generalization, but if a model is well-calibrated on a held-out set, its confidence may provide some signal for distinguishing OoD examples, so we mention it as a relevant metric for \textbf{(RG3)}.

To our knowledge, the only works to date that have demonstrated progress on robust generalization for modern machine learning datasets are the \textit{Variational Information Bottleneck}~\citep{vib,uncertainvib} (VIB), and \textit{Information Dropout}~\citep{achille2018information}.
\citet{vib} presented preliminary results that VIB improves adversarial robustness on image classification tasks while maintaining high classification accuracy (\textbf{(RG1)} and \textbf{(RG2)}).
\citet{uncertainvib} showed that VIB models provide a useful signal, the \textit{Rate}, $R$, for detecting OoD examples (\textbf{(RG3)}).
\citet{achille2018information} also showed preliminary results on adversarial robustness and demonstrated failure to train on information-free datasets.

In this work, we do not claim to ``solve'' robust generalization, but we do show notable improvement on all three conditions simply by changing the training objective.
This evidence supports our core hypothesis that problems of robust generalization are caused in part by retaining too much information about the training data.

\section{The Minimum Necessary Information}
\label{sec:mni}

We define the \textit{Minimum Necessary Information} (MNI) criterion for a learned representation in three parts:
\begin{itemize}[leftmargin=1em]
  \item \textbf{Information.}
    We would like a representation $Z$ that captures useful information about a dataset $(X,Y)$.
    Entropy is the unique measure of information~\citep{shannon}\footnote{%
      We assume familiarity with the mutual information and its relationships to entropy and conditional entropy: $I(X;Y) = H(X) - H(X|Y) = H(Y) - H(Y|X)$~\citep{infobook}.
    }, so the criterion prefers information-theoretic approaches.
  \item \textbf{Necessity.}%
    \footnote{%
      \citet{achille2018information} and other authors call this \textit{sufficiency}.
      We avoid the term due to potential confusion with minimum sufficient statistics, which maintain the mutual information between a model and the data it generates.
    }
    The semantic value of information is given by a \textit{task}, which is specified by the set of variables in the dataset.
    Here we will assume that the task of interest is to predict $Y$ given $X$ using our representation $Z$.
  \item \textbf{Minimality.}
    Given all representations $\mathcal{Z}$ that can solve the task, we require one that retains the smallest amount of information about the task: $\inf_{Z \in \mathcal{Z}} I(Z;X,Y)$.
\end{itemize}

\textit{Necessity} can be defined as $I(X;Y) \le I(Y;Z)$.
Any less information than that would prevent $Z$ from solving the task of predicting $Y$ from $X$.
\textit{Minimality} can be defined as $I(X;Y) \ge I(X;Z)$.
Any more information than that would result in $Z$ having redundant information about $X$ that is unnecessary for predicting $Y$.
Since $Z$ is constrained from above and below, we have the following necessary and sufficient conditions for perfectly achieving the Minimum Necessary Information, which we call the \textit{MNI Point}:
\begin{align}\label{eqn:mni_info}
I(X;Y) = I(X;Z) = I(Y;Z)
\end{align}

In general it may not be possible to satisfy~\Cref{eqn:mni_info}.
As discussed in~\citet{anantharam2013maximal,strongdpi,wu2019learnability}, for any given dataset $(X,Y)$, there is some maximum value for any possible representation $Z$:
\begin{equation}
1 \ge \eta_{\operatorname{KL}} = \sup_{Z \leftarrow X \rightarrow Y} \frac{I(Y;Z)}{I(X;Z)}
\end{equation}
with equality only when $X \rightarrow Y$ is a \textit{deterministic} map.
Training datasets are often deterministic in one direction or the other.
E.g., common image datasets map each distinct image to a single label.
Thus, in practice, we can often get very close to the MNI given a sufficiently powerful model.

\paragraph{MNI and robust generalization.}
To satisfy \textbf{(RG1)} (classical generalization), a model must have $I(X;Z) \ge I(X;Y) = I(Y;Z)$ on the \emph{test} dataset.
\citet{shamir2010learning} show that $|I(X;Z) - \hat{I}(X;Z)| \approx O\Big(\frac{2^{\hat{I}(X;Z)}}{\sqrt{N}}\Big)$, where $\hat{I}(\cdot)$ indicates the training set information and $N$ is the size of the training set.
More recently, \citet{bassily2017learners} gave a similar result in a PAC setting.
Both results indicate that models that are \emph{compressed on the training data} should do \emph{better at generalizing} to similar test data.

Less clear is how an MNI model might improve on \textbf{(RG2)} (adversarial robustness).
In this work, we treat it as a hypothesis that we investigate empirically rather than theoretically.
The intuition behind the hypothesis can be described in terms of the idea of \textit{robust} and \textit{non-robust features} from~\citet{ilyas2019adversarial}: non-robust features in $X$ should be compressed as much as possible when we learn $Z$, whereas robust features should be retained as much as is necessary.
If~\Cref{eqn:mni_info} is satisfied, $Z$ must have ``scaled'' the importance of the the features in $X$ according to their importance for predicting $Y$.
Consequently, an attacker that tries to take advantage of a non-robust feature will have to change it much more in order to confuse the model, possibly exceeding the constraints of the attack before it succeeds.

For \textbf{(RG3)} (detection), the MNI criterion does not directly apply, as that will be a property of specific modeling choices.
However, if the model provides an accurate way to measure $I(X=x;Z=z)$ for a particular pair $(x,z)$, \citet{uncertainvib} suggests that can be a valuable signal for OoD detection.

\section{The Conditional Entropy Bottleneck}
\label{sec:ceb}

We would like to learn a representation $Z$ of $X$ that will be useful for predicting $Y$.
We can represent this problem setting with the Markov chain $Z \leftarrow X \leftrightarrow Y$.
We would like $Z$ to satisfy~\Cref{eqn:mni_info}.
Given the conditional independence $Z \Perp Y | X$ in our Markov chain, $I(Y;Z) \le I(X;Y)$, by the data processing inequality.
Thus, maximizing $I(Y;Z)$ is consistent with the MNI criterion.

However, $I(X;Z)$ does not clearly have a constraint that targets $I(X;Y)$, as $0 \le I(X;Z) \le H(X)$.
Instead, we can notice the following identities at the MNI point:
\begin{equation}
  I(X;Y|Z) = I(X;Z|Y) = I(Y;Z|X) = 0
\end{equation}
The conditional mutual information is always non-negative, so learning a compressed representation $Z$ of $X$ is equivalent to minimizing $I(X;Z|Y)$.
Using our Markov chain and the chain rule of mutual information~\citep{infobook}:
\begin{equation}
  I(X;Z|Y) = I(X,Y;Z) - I(Y;Z) = I(X;Z) - I(Y;Z)
  \label{eqn:condinfo}
\end{equation}
This leads us to the general \textit{Conditional Entropy Bottleneck}:
\begin{align}
\label{eqn:ceb}
\operatorname{CEB} &\equiv \min_Z I(X;Z|Y) - \gamma I(Y;Z) \\
&= \min_Z H(Z) - H(Z|X) - H(Z) + H(Z|Y) \nonumber \\
&\quad \quad \quad - \gamma (H(Y) + H(Y|Z)) \\
&\Leftrightarrow \min_Z -H(Z|X) + H(Z|Y) + \gamma H(Y|Z) \label{eqn:cebnohy}
\end{align}
In line~\ref{eqn:cebnohy}, we can optionally drop $H(Y)$ because it is constant with respect to $Z$.
Here, any $\gamma > 0$ is valid, but for deterministic datasets (\Cref{sec:mni}), $\gamma = 1$ will achieve the MNI for a sufficiently powerful model.
Further, we should expect $\gamma=1$ to yield \emph{consistent} models and other values of $\gamma$ not to:
  since $I(Y;Z)$ shows up in two forms in the objective, weighing them differently forces the optimization procedure to count bits of $I(Y;Z)$ in two different ways, potentially leading to a situation where $H(Z) - H(Z|Y) \ne H(Y) - H(Y|Z)$ at convergence.
Given knowledge of those four entropies, we can define a consistency metric for $Z$:
\begin{align}
\label{eqn:consistency}
C_{I(Y;Z)}(Z) \equiv |H(Z) - H(Z|Y) - H(Y) + H(Y|Z)|
\end{align}

\subsection{Variational Bound on CEB}

We will variationally upper bound the first term of~\Cref{eqn:ceb} and lower bound the second term using three distributions:
$e(z|x)$, the \textit{encoder} which defines the joint distribution we will use for sampling, $p(x,y,z) \equiv p(x,y) e(z|x)$;
$b(z|y)$, the \textit{backward encoder}, an approximation of $p(z|y)$;
and $c(y|z)$, the \textit{classifier}, an approximation of $p(y|z)$ (the name is arbitrary, as $Y$ may not be labels).
All of $e(\cdot)$, $b(\cdot)$, and $c(\cdot)$ may have learned parameters, just like the encoder and decoder of a VAE~\citep{vae}, or the encoder, classifier, and marginal in VIB.

The first term of~\Cref{eqn:ceb}:\footnote{
  We write expectations $\left\langle \log e(z|x) \right\rangle$.
  They are always with respect to the joint distribution; here, that is $p(x,y,z) \equiv p(x,y) e(z|x)$.
}
\begin{align}
I(X;Z|Y) =& \, {-H}(Z|X) + H(Z|Y) = \left\langle \log e(z|x) \right\rangle - \left\langle \log p(z|y) \right\rangle \\
=& \left\langle \log e(z|x) \right\rangle - \left\langle \log b(z|y) \right\rangle -  \left\langle \operatorname{KL}[p(z|y)||b(z|y)] \right\rangle \\
\le& \left\langle \log e(z|x) \right\rangle - \left\langle \log b(z|y) \right\rangle
\end{align}

The second term of~\Cref{eqn:ceb}:
\begin{align}
I(Y;Z) =& \, H(Y) - H(Y|Z) \propto {-H}(Y|Z) = \left\langle \log p(y|z) \right\rangle \\
=& \left\langle \log c(y|z) \right\rangle + \left\langle \operatorname{KL}[p(y|z)||c(y|z)]  \right\rangle \\
\ge& \left\langle \log c(y|z) \right\rangle
\end{align}

These variational bounds give us a tractable objective function for amortized inference, the \textit{Variational Conditional Entropy Bottleneck} (VCEB):
\begin{align}
\operatorname{VCEB} \equiv& \min_{e,b,c}\left\langle \log e(z|x) \right\rangle - \left\langle \log b(z|y) \right\rangle - \gamma \left\langle \log c(y|z) \right\rangle
\end{align}

There are a number of other ways to optimize~\Cref{eqn:ceb}.
We describe a few of them in~\Cref{sec:models} and the supplementary material.

\begin{figure}[t]
  \center
  \begin{tikzpicture}[scale=1.5]
    \begin{scope}
      \clip ( 0.5, 0.0) circle (0.7cm);
      \fill[pattern=crosshatch, pattern color=red!80!black] ( 0.5, 0.0) circle (0.7cm);
      \clip (-0.5, 0.0) circle (0.7cm);
      \fill[white] ( 0.5, 0.0) circle (0.7cm);
    \end{scope}
    \begin{scope}
      \fill[pattern=horizontal lines, pattern color=cyan!80!black] (-0.5, 0.0) circle (0.7cm);
    \end{scope}
    \begin{scope}
      \clip (-0.5, 0.0) circle (0.7cm);
      \clip ( 0.5, 0.0) circle (0.7cm);
      \fill[pattern=vertical lines, pattern color=orange!80!black] ( 0.5, 0.0) circle (0.7cm);
    \end{scope}
    \draw (-0.5, 0.0) circle (0.7cm) +(0.0, +0.85) node{$H(X)$};
    \draw ( 0.5, 0.0) circle (0.7cm) +(0.0, +0.85) node{$H(Y)$};
  \end{tikzpicture}
  \quad
  \begin{tikzpicture}[scale=1.5]
    \begin{scope}
      \fill[pattern=crosshatch, pattern color=red!80!black] ( 0.5, 0.0) circle (0.7cm);
      \clip (-0.5, 0.0) circle (0.7cm);
      \fill[white] ( 0.5, 0.0) circle (0.7cm);
    \end{scope}
    \begin{scope}
      \fill[pattern=horizontal lines, pattern color=cyan!80!black] (-0.5, 0.0) circle (0.7cm);
      \clip ( 0.5, 0.0) circle (0.7cm);
      \fill[white] (-0.5, 0.0) circle (0.7cm);
    \end{scope}
    \begin{scope}
      \clip (-0.5, 0.0) circle (0.7cm);
      \clip ( 0.5, 0.0) circle (0.7cm);
      \fill[pattern=vertical lines, pattern color=orange!80!black] ( 0.5, 0.0) circle (0.7cm);
    \end{scope}
    \draw (-0.5, 0.0) circle (0.7cm) +(0.0, +0.85) node{$H(X)$};
    \draw ( 0.5, 0.0) circle (0.7cm) +(0.0, +0.85) node{$H(Y)$};
  \end{tikzpicture}

  \vspace{0.2cm}

  \textbf{~~~ IB}
  \quad \quad \quad \quad \quad \quad \quad \quad \quad \quad
  \textbf{CEB}

  \caption{%
    Information diagrams showing how IB and CEB maximize and minimize different regions.
\raisebox{0ex}[0ex][0ex]{
\tikz[baseline={([yshift=-0.5ex]current bounding box.center)}]{ \draw[pattern=crosshatch, pattern color=red!80!black] ( 0.0, 0.0) circle (0.5em) }
}\!:
    regions inaccessible to the objective due to the Markov chain  $Z \leftarrow X \leftrightarrow Y$.
\raisebox{0ex}[0ex][0ex]{
\tikz[baseline={([yshift=-0.5ex]current bounding box.center)}]{ \draw[pattern=vertical lines, pattern color=orange!80!black] ( 0.0, 0.0) circle (0.5em) }
}\!:
    regions being maximized by the objective ($I(Y;Z)$ in both cases).
\raisebox{0ex}[0ex][0ex]{
\tikz[baseline={([yshift=-0.5ex]current bounding box.center)}]{ \draw[pattern=horizontal lines, pattern color=cyan!80!black] ( 0.0, 0.0) circle (0.5em); }
}\!:
    regions being minimized by the objective.
    \textbf{IB} minimizes the intersection between $Z$ and both $H(X|Y)$ and $I(X;Y)$.
    \textbf{CEB} only minimizes the intersection between $Z$ and $H(X|Y)$.
  }
  \label{fig:xyvenn}
\end{figure}

\begin{figure}[th]
  \centering
  \includegraphics[width=0.585\linewidth]{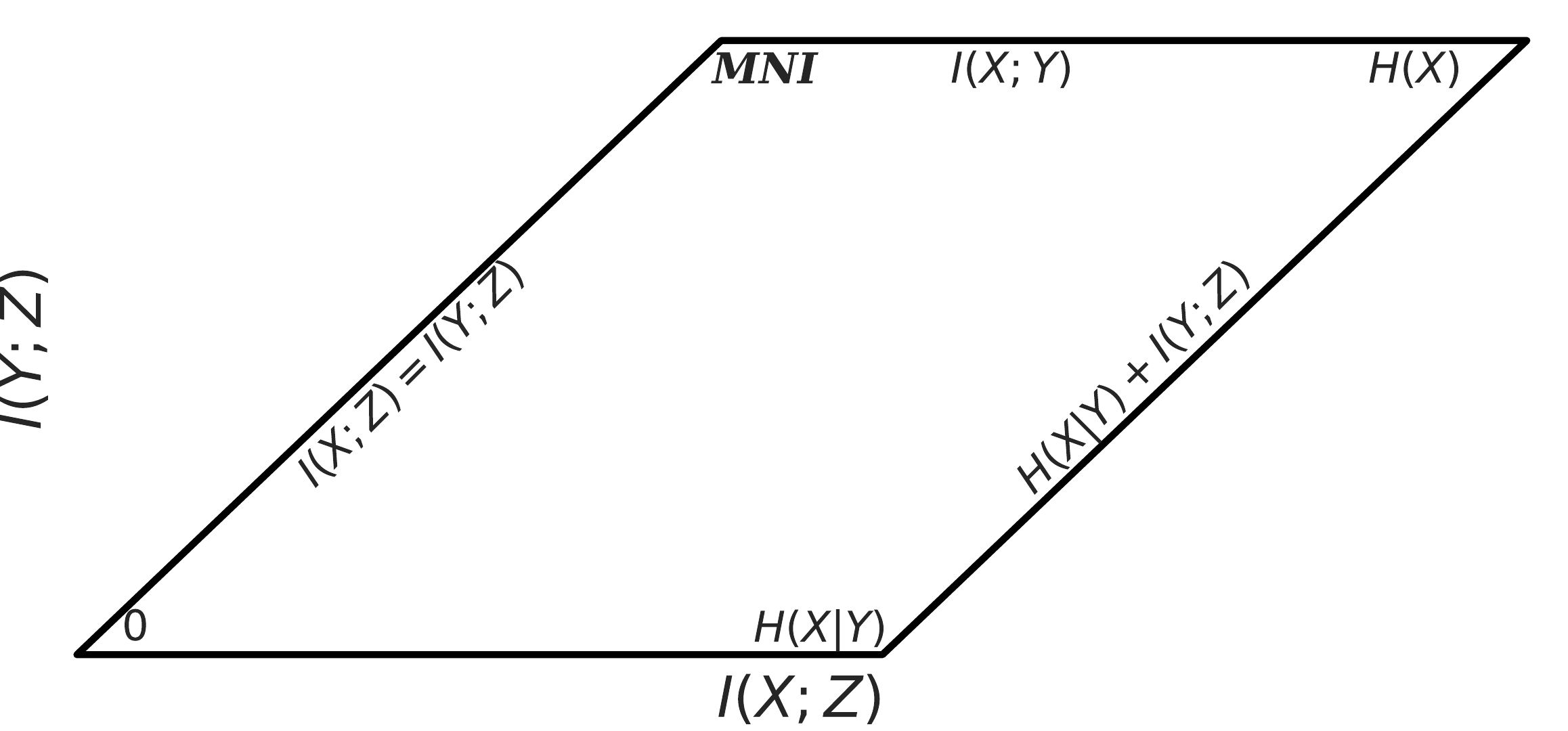}
  \quad
  \includegraphics[width=0.338\linewidth]{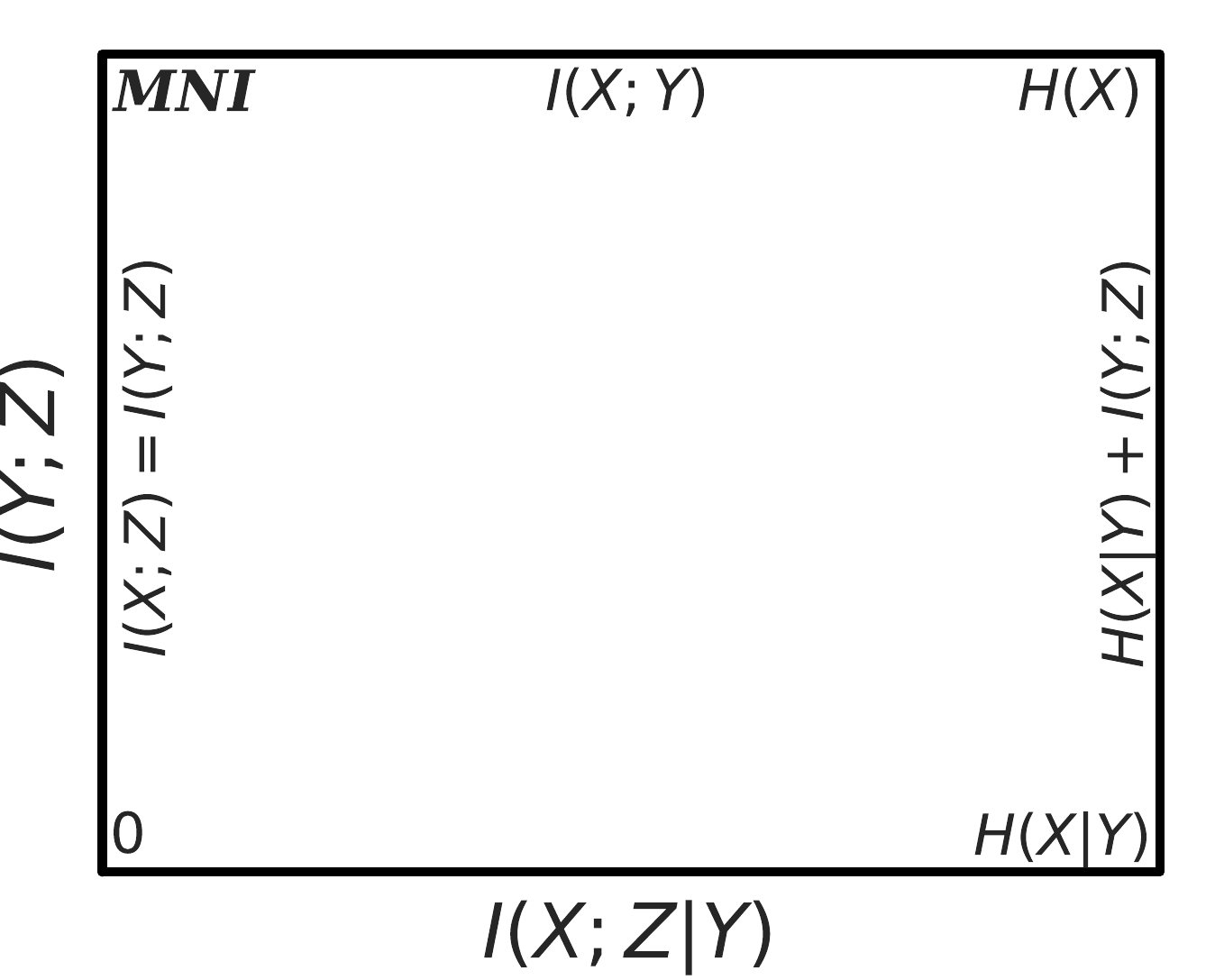}

  \vspace{0.2cm}

  \textbf{\quad \quad \quad \quad IB}
  \quad \quad \quad \quad \quad \quad \quad \quad \quad \quad \quad
  \textbf{CEB}

  \caption{
    Geometry of the feasible regions for IB and CEB, with all points labeled.
    CEB rectifies IB's parallelogram by subtracting $I(Y;Z)$ at every point.
    Everything outside of the black lines is unattainable by any model on any dataset.
  }
  \label{fig:opt_geometry}
\end{figure}

\section{Comparison to the Information Bottleneck}
\label{sec:vib}

The Information Bottleneck (IB)~\citep{tishby2000information} learns a representation $Z$ from $X$ subject to a soft constraint:
\begin{equation}
  \label{eqn:ib}
  IB \equiv \min_Z I(X;Z) - \beta I(Y;Z)
\end{equation}
where $\beta^{-1}$ controls the strength of the constraint.
As $\beta \rightarrow \infty$, IB recovers the standard cross-entropy loss.

In~\Cref{fig:xyvenn} we show information diagrams comparing which regions IB and CEB maximize and minimize.\footnote{%
  See~\citet{yeung1991new} for a theoretical explanation of information diagrams.
}
CEB avoids trying to both minimize and maximize the central region at the same time.
In~\Cref{fig:opt_geometry} we show the feasible regions for CEB and IB, labeling the MNI point on both.
CEB's rectification of the information plane means that we can always measure in absolute terms how much more we could compress our representation \emph{at the same predictive performance}: $I(X;Z|Y) \ge 0$.
For IB, it is not possible to tell \textit{a priori} how far we are from optimal compression.

From~\Cref{eqn:condinfo,eqn:ceb,eqn:ib}, it is clear that CEB and IB are equivalent for $\gamma = \beta - 1$.
To simplify comparison of the two objectives, we can parameterize them with:
\begin{equation}
  \rho = \log \gamma = \log(\beta - 1)
\end{equation}
Under this parameterization, for deterministic datasets, sufficiently powerful models will target the MNI point at $\rho=0$.
As $\rho$ increases, more information is captured by the model.
$\rho < 0$ \emph{may} capture less than the MNI.
$\rho > 0$ \emph{may} capture more than the MNI.

\paragraph{Amortized IB.}
As described in~\citet{tishby2000information}, IB is a tabular method, so it is not usable for amortized inference.\footnote{%
  The tabular optimization procedure used for IB trivially applies to CEB, just by setting $\beta=2$.
}
Two recent works have extended IB for amortized inference.
\citet{achille2018information} presents \textit{InfoDropout}, which uses IB to motivate a variation on Dropout~\citep{dropout}.
\citet{vib} presents the \textit{Variational Information Bottleneck} (VIB):
\begin{equation}
VIB \equiv \left\langle \log e(z|x) \right\rangle - \left\langle \log m(z) \right\rangle) - \beta \left\langle \log c(y|z) \right\rangle
\end{equation}
Instead of the backward encoder, VIB has a \textit{marginal posterior}, $m(z)$, which is a variational approximation to $e(z)=\int dx\, p(x) e(z|x)$.

Following~\citet{brokenelbo}, we define the \textit{Rate} ($R$):
\begin{align}
R \equiv \left\langle \log e(z|x) \right\rangle - \left\langle \log m(z) \right\rangle \ge I(X;Z)
\end{align}
We similarly define the \textit{Residual Information} ($Re_X$):
\begin{align}
Re_X \equiv \left\langle \log e(z|x) \right\rangle - \left\langle \log b(z|y) \right\rangle \ge I(X;Z|Y)
\end{align}
During optimization, observing $R$ does not tell us how tightly we are adhering to the MNI.
However, observing $Re_X$ tells us exactly how many bits we are from the MNI point, assuming that our current classifier is optimal.

For convenience, define $CEB_x \equiv CEB_{\rho=x}$, and likewise for VIB.
We can compare variational CEB with VIB by taking their difference at $\rho=0$:
\begin{align}
  CEB_0 - VIB_0 =& \left\langle \log b(z|y) \right\rangle - \left\langle \log m(z) \right\rangle \\
  &~ - \left\langle \log c(y|z) \right\rangle + \left\langle \log p(y) \right\rangle
\end{align}

Solving for $m(z)$ when that difference is 0:
\begin{equation}
  m(z) = \frac{b(z|y)p(y)}{c(y|z)}
\end{equation}

Since the optimal $m^{\ast}(z)$ is the marginalization of $e(z|x)$, at convergence we must have:
\begin{equation}
  m^{\ast}(z) = \int dx\, p(x) e(z|x) = \frac{p(z|y)p(y)}{p(y|z)}
\end{equation}

This solution may be difficult to find, as $m(z)$ only gets information about $y$ indirectly through $e(z|x)$.
For otherwise equivalent models, we may expect $VIB_0$ to converge to a looser approximation of $I(X;Z) = I(Y;Z) = I(X;Y)$ than CEB.
Since VIB optimizes an upper bound on $I(X;Z)$, $VIB_0$ will report $R$ converging to $I(X;Y)$, but may capture less than the MNI.
In contrast, if $Re_X$ converges to 0, the variational tightness of $b(z|y)$ to the optimal $p(z|y)$ depends only on the tightness of $c(y|z)$ to the optimal $p(y|z)$.

\section{Model Variants}
\label{sec:models}

We introduce some variants on the basic variational CEB classification model that we will use in~\Cref{sec:cifar}.

\subsection{Bidirectional CEB}
\label{sec:bidir}

We can learn a shared representation $Z$ that can be used to predict both $X$ and $Y$ with the following bidirectional CEB model: $Z_X \leftarrow X \leftrightarrow Y \rightarrow Z_Y$.
This corresponds to the following joint: $p(x,y,z_X,z_Y) \equiv p(x,y) e(z_X|x) b(z_Y|y)$.
The main CEB objective can then be applied in both directions:
\begin{align}
\text{CEB}_{\text{bidir}} \equiv \min&\,\,{-H}(Z_X|X) + H(Z_X|Y) + \gamma_X H(Y|Z_X) \nonumber \\
&- H(Z_Y|Y) + H(Z_Y|X) + \gamma_Y H(X|Z_Y)
\end{align}

For the two latent representations to be useful, we want them to be consistent with each other (minimally, they must have the same parametric form).
Fortunately, that consistency is trivial to encourage by making the natural variational substitutions: $p(z_Y|x) \rightarrow e(z_Y|x)$ and $p(z_X|y) \rightarrow b(z_X|y)$.
This gives variational CEB$_{\text{bidir}}$:
\begin{align}
\min &\left\langle \log e(z_X|x) \right\rangle - \left\langle \log b(z_X|y) \right\rangle - \gamma_X \left\langle \log c(y|z_X) \right\rangle \nonumber \\
   + &\left\langle \log b(z_Y|y) \right\rangle - \left\langle \log e(z_Y|x) \right\rangle - \gamma_Y \left\langle \log d(x|z_Y) \right\rangle
\end{align}
where $d(x|z)$ is a \textit{decoder} distribution.
At convergence, we learn a unified $Z$ that is consistent with both $Z_X$ and $Z_Y$, permitting generation of either output given either input in the trained model, in the same spirit as~\citet{imagination}, but without needing to train a joint encoder $q(z|x,y)$.

\subsection{Consistent Classifier}

We can reuse the backwards encoder as a classifier: $c(y|z) \propto b(z|y) p(y)$.
We refer to this as the \textit{Consistent Classifier}: $c(y|z) \equiv \operatorname{softmax} b(z|y) p(y)$.
If the labels are uniformly distributed, the $p(y)$ factor can be dropped; otherwise, it suffices to use the empirical $p(y)$.
Using the consistent classifier for classification problems results in a model that only needs parameters for the two encoders, $e(z|x)$ and $b(z|y)$.
This classifier differs from the more common \textit{maximum a posteriori} (MAP) classifier because $b(z|y)$ is not the sampling distribution of either $Z$ or $Y$.

\subsection{CatGen Decoder}

We can further generalize the idea of the consistent classifier to arbitrary prediction tasks by relaxing the requirement that we perfectly marginalize $Y$ in the softmax.
Instead, we can marginalize $Y$ over any minibatch of size $K$ we see at training time, under an assumption of a uniform distribution over the training examples we sampled:
\begin{align}
p(y|z) =& \frac{p(z|y)p(y)}{\int dy'\, p(z|y')p(y')} \\
\approx& \frac{p(z|y) \frac 1 K}{\sum_{k=1}^K p(z|y_k) \frac 1 K}
= \frac{p(z|y)}{\sum_{k=1}^K p(z|y_k)} \\
\approx& \frac{b(z|y)}{\sum_{k=1}^K b(z|y_k)} \equiv c(y|z)
\end{align}
We can immediately see that this definition of $c(y|z)$ gives a valid distribution, as it is just a softmax over the minibatch.
That means it can be directly used in the original objective without violating the variational bound.
We call this decoder \textit{CatGen}, for \textit{Categorical Generative Model} because it can trivially ``generate'' $Y$: the softmax defines a categorical distribution over the batch; sampling from it gives indices of $Y=y_j$ that most closely correspond to $Z=z_i$.

Maximizing $I(Y;Z)$ in this manner is a universal task, in that it can be applied to any paired data $X,Y$.
This includes images and labels -- the CatGen model may be used in place of both $c(y|z_X)$ and $d(x|z_Y)$ in the CEB$_{\text{bidir}}$ model (using $e(z|x)$ for $d(x|z_Y)$).
This avoids a common concern when dealing with multivariate predictions: if predicting $X$ is disproportionately harder than predicting $Y$, it can be difficult to balance the model~\citep{imagination,higgins2018scan}.
For CatGen models, predicting $X$ is never any harder than predicting $Y$, since in both cases we are just trying to choose the correct example out of $K$ possibilities.

It turns out that CatGen is mathematically equivalent to \textit{Contrastive Predictive Coding} (CPC)~\citep{oord2018representation} after an offset of $\log K$.
We can see this using the proof from~\citet{vmibounds}, and substituting $\log b(z|y)$ for $f(y,z)$:
\begin{align}
I(Y;Z) &\le \frac 1 K \sum_{k=1}^{K} \mathbb{E}_{\prod_{j} y_k,z \sim p(y_j) p(x_k|y_k) e(z|x_k)} \Bigg[ \log \frac{e^{f(y_k, z)}}{\frac 1 K \sum_{i=1}^{K} e^{f(y_i, z)}} \Bigg] \\
&= \frac 1 K \sum_{k=1}^{K} \mathbb{E}_{\prod_{j} y_k,z \sim p(y_j) p(x_k|y_k) e(z|x_k)} \Bigg[ \log \frac{b(z|y_k)}{\frac 1 K \sum_{i=1}^{K} b(z|y_i)} \Bigg]
\end{align}
The advantage of the CatGen approach over CPC in the CEB setting is that we already have parameterized the forward and backward encoders to compute $I(X;Z|Y)$, so we don't need to introduce any new parameters when using CatGen to maximize the $I(Y;Z)$ term.

As with CPC, the CatGen bound is constrained by $\log K$, but when targeting the MNI, it is more likely that we can train with $\log K \ge I(X;Y)$.
This is trivially the case for the datasets we explore here, where $I(X;Y) \le \log 10$.
It is also practical for larger datasets like ImageNet, where models are routinely trained with batch sizes in the thousands (e.g., \citet{goyal2017accurate}), and $I(X;Y) \le \log 1000$.

\section{Experiments}

We evaluate deterministic, VIB, and CEB models on Fashion MNIST~\citep{fashion} and CIFAR10~\citep{cifar}.
Our experiments focus on comparing the performance of otherwise \emph{identical} models when we change only the objective function and vary $\rho$.
Thus, we are interested in relative differences in performance that can be directly attributed to the difference in objective and $\rho$.
These experiments cover the three aspects of \textit{Robust Generalization} (\Cref{sec:robust}):
\textbf{(RG1)} (classical generalization) in \Cref{sec:fashion,sec:cifar};
\textbf{(RG2)} (adversarial robustness) in \Cref{sec:fashion,sec:cifar};
and \textbf{(RG3)} (detection) in \Cref{sec:fashion}.

\subsection{(RG1), (RG2), and (RG3): Fashion MNIST}
\label{sec:fashion}

\begin{figure}[htb]
\centering
\vspace{-0.325cm}
\includegraphics[width=0.7\linewidth,trim={0 0 9.55cm 0},clip]{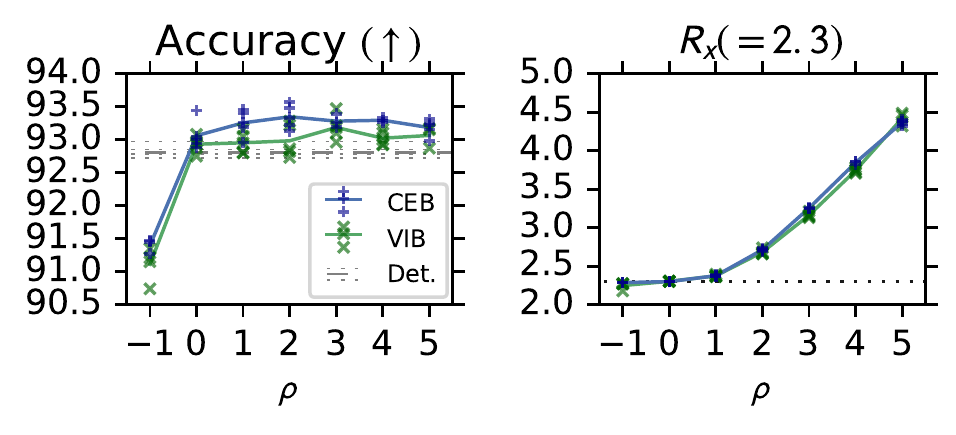}
\vspace{-0.7cm}

\includegraphics[width=0.7\linewidth,trim={9.55cm 0 0 0},clip]{figs/fashion_targeted_plots.pdf}
\vspace{-0.5cm}
\caption{%
  Test accuracy, maximum rate lower bound $R_X \le I(Z;X)$ seen during training, and robustness to targeted PGD L$_2$ and L$_\infty$ attacks on CEB, VIB, and Deterministic models trained on Fashion MNIST.
  At every $\rho$ the CEB models outperform the VIB models on both accuracy and robustness, while having essentially identical maximum rates.
  \emph{None of these models is adversarially trained.}
}
\label{fig:fashion_targeted_plots}
\end{figure}

Fashion MNIST~\citep{fashionmnist} is an interesting dataset in that it is visually complex and challenging, but small enough to train in a reasonable amount of time.
We trained 60 different models on Fashion MNIST, four each for the following 15 types:
a deterministic model (\textit{Determ});
seven VIB models (VIB$_{-1}$, ..., VIB$_5$); and seven CEB models (CEB$_{-1}$, ..., CEB$_5$).
Subscripts indicate $\rho$.
All 60 models share the same inference architecture and are trained with otherwise identical hyperparameters.\footnote{%
  See the supplemental material for details.
}

\paragraph{(RG1): Accuracy and compression.}
\label{sec:classification}
In~\Cref{fig:fashion_targeted_plots} we see that both VIB and
CEB have improved accuracy over the deterministic baseline, consistent with compressed representations generalizing better.
Also, CEB outperforms VIB at every $\rho$, which we can attribute to the tighter variational bound given by minimizing $Re_X$ rather than $R$.
In the case of a simple classification problem with a uniform distribution over classes in the training set (like Fashion MNIST), we can directly compute $I(X;Y) = \log C$, where $C$ is the number of classes.
In order to compare the relative complexity of the learned representations for the VIB and CEB models,
in the second panel of~\Cref{fig:fashion_targeted_plots} we show the maximum \textit{rate lower bound} seen during training: $R_X \equiv \left \langle \log \frac{e(z|x)}{\frac 1 K \sum_k^K e(z|x_k)} \right \rangle \le I(X;Z)$ using the encoder's minibatch marginal for both VIB and CEB.\footnote{%
  This lower bound on $I(X;Z)$ is the ``InfoNCE with a tractable encoder'' bound from~\citet{vmibounds}.
}
The two sets of models show nearly the same $R_X$ at each value of $\rho$.
Both models converge to exactly $I(X;Y) = \log 10 \approx 2.3$ nats at $\rho=0$, as predicted by the derivation of CEB.

\paragraph{(RG2): Adversarial robustness.}
\label{sec:adversarial}
The bottom two panels of~\Cref{fig:fashion_targeted_plots} show robustness to targeted \textit{Projected Gradient Descent} (PGD) L$_2$ and L$_\infty$ attacks~\citep{madry2017towards}.
All of the attacks are targeting the \textit{trouser} class of Fashion MNIST, as that is the most distinctive class.
Targeting a less distinctive class, such as one of the shirt classes, would confuse the difficulty of classifying the different shirts and the robustness of the model to adversaries.
To measure robustness to the targeted attacks, we count the number of predictions that changed from a correct prediction on the clean image to an incorrect prediction of the target class on the adversarial image, and divide by the original number of correct predictions.
Consistent with testing \textbf{(RG2)}, these adversaries are completely unknown to the models at training time -- none of these models see any adversarial examples during training.
CEB again outperforms VIB at every $\rho$, and the deterministic baseline at all but the least-compressed model ($\rho=5$).
We also see for both models that as $\rho$ decreases, the robustness to both attacks increases, indicating that more compressed models are more robust.

Consistent with the MNI hypothesis, at $\rho=0$ we end up with CEB models that have hit exactly 2.3 nats for the rate lower bound, have maintained high accuracy, and have strong robustness to both attacks.
Moving to $\rho=-1$ gives only a small improvement to robustness, at the cost of a large decrease in accuracy.

\begin{table}[tb]
  \centering
  \small

  \setlength\tabcolsep{3.6pt} 
  \begin{tabular}{|c|c|c|c|c|c|}
  \hline

  OoD & \makecell{Model \\ Adv Succ $\downarrow$} & Thrsh. & \makecell{FPR @\\95\% TPR $\downarrow$} & \makecell{AUROC \\ $\uparrow$} & \makecell{AUPR \\ In $\uparrow$} \\
  \hhline{|=|=|=|=|=|=|}

  \multirow{4}{*}{U(0,1)}

    & \multirow{1}{*}{Determ}
      & $H$ & 35.8 & 93.5 & 97.1 \\

    \cline{2-6}

    & \multirow{1}{*}{VIB$_4$}  
      & $R$ & \textbf{0.0} & \textbf{100.0} & \textbf{100.0} \\

    \cline{2-6}

    & \multirow{1}{*}{VIB$_0$}
      & $R$ & 80.6 & 57.1 & 51.4 \\

    \cline{2-6}

    & \multirow{1}{*}{CEB$_0$}  
      & $R$ & \textbf{0.0} & \textbf{100.0} & \textbf{100.0} \\

  \hhline{|=|=|=|=|=|=|}

  \multirow{4}{*}{MNIST}

    & \multirow{1}{*}{Determ}
      & $H$ & 59.0 & 88.4 & 90.0 \\

    \cline{2-6}

    & \multirow{1}{*}{VIB$_4$}
      & $R$ & \textbf{0.0} & \textbf{100.0} & \textbf{100.0} \\

    \cline{2-6}

    & \multirow{1}{*}{VIB$_0$}
      & $R$ & 12.3 & 66.7 & 91.1 \\

    \cline{2-6}

    & \multirow{1}{*}{CEB$_0$}  
      & $R$ & \textbf{0.1} & 94.4 & \textbf{99.9} \\

  \hhline{|=|=|=|=|=|=|}

  \multirow{4}{*}{\makecell{Vertical \\ Flip}}

    & \multirow{1}{*}{Determ}
      & $H$ & 66.8 & 88.6 & 90.2 \\

    \cline{2-6}

    & \multirow{1}{*}{VIB$_4$}
      & $R$ & \textbf{0.0} & \textbf{100.0} & \textbf{100.0} \\

    \cline{2-6}

    & \multirow{1}{*}{VIB$_0$}
      & $R$ & 17.3 & 52.7 & 91.3 \\

    \cline{2-6}

    & \multirow{1}{*}{CEB$_0$}  
      & $R$ & \textbf{0.0} & 90.7 & \textbf{100.0} \\
    \cline{2-6}

  \hhline{|=|=|=|=|=|=|}

  \multirow{8}{*}{CW}

    & Determ
        & \multirow{2}{*}{$H$} & \multirow{2}{*}{15.4} & \multirow{2}{*}{90.7} & \multirow{2}{*}{86.0} \\
    & $100.0\%$ & & & & \\

    \cline{2-6}

    & VIB$_4$
        & \multirow{2}{*}{$R$} & \multirow{2}{*}{\textbf{0.0}} & \multirow{2}{*}{\textbf{100.0}} & \multirow{2}{*}{\textbf{100.0}} \\
    & $55.2\%$ & & & &
    \\

    \cline{2-6}

    & VIB$_0$
        & \multirow{2}{*}{$R$} & \multirow{2}{*}{\textbf{0.0}} & \multirow{2}{*}{98.7} & \multirow{2}{*}{\textbf{100.0}} \\
    & \textbf{35.8\%} & & & &
    \\

    \cline{2-6}

    & CEB$_0$
        & \multirow{2}{*}{$R$} & \multirow{2}{*}{\textbf{0.0}} & \multirow{2}{*}{\textbf{99.7}} & \multirow{2}{*}{\textbf{100.0}} \\
    & \textbf{35.8\%} & & & &
    \\

    \hline
  \end{tabular}

  \caption{
    Results for out-of-distribution detection (\textit{OoD}).
    \textit{Thrsh.} is the threshold score used: $H$ is the entropy of the classifier; $R$ is the rate.
    Determ cannot compute $R$, so only $H$ is shown.
    For VIB and CEB models, $H$ is always inferior to $R$, similar to findings in~\citet{uncertainvib}, so we omit it.
    \textit{Adv Succ} is attack success of the CW adversary (bottom rows).
    Arrows denote whether higher or lower scores are better.
    \textbf{Bold} indicates the best score in that column for that OoD dataset.
  }
  \label{tab:ood}
\end{table}

\begin{figure}[t]
  \centering
  \raisebox{0.12\linewidth}{\textbf{a}}
  \includegraphics[width=0.25\linewidth]{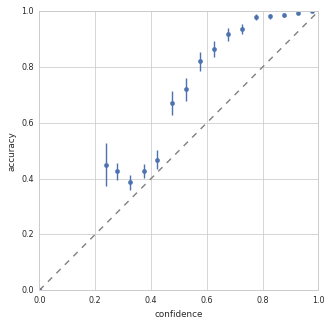}
  \includegraphics[width=0.25\linewidth]{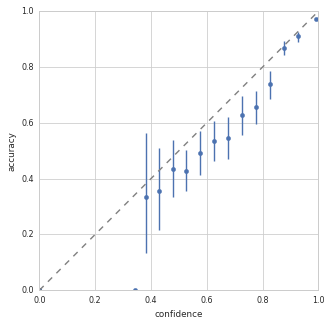}
  \includegraphics[width=0.25\linewidth]{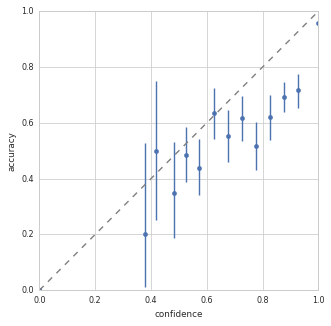} \\
  \raisebox{0.12\linewidth}{\textbf{b}}
  \includegraphics[width=0.25\linewidth]{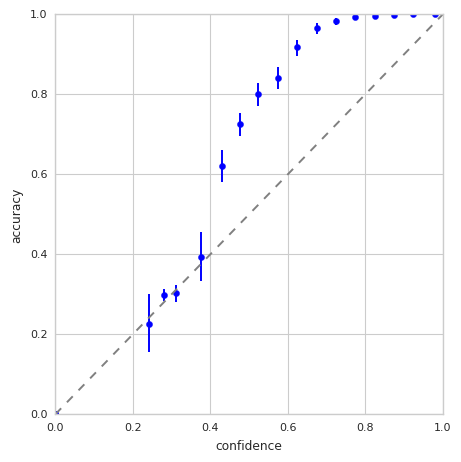}
  \includegraphics[width=0.25\linewidth]{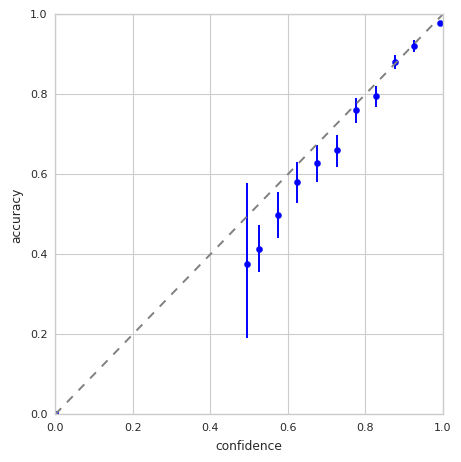}
  \includegraphics[width=0.25\linewidth]{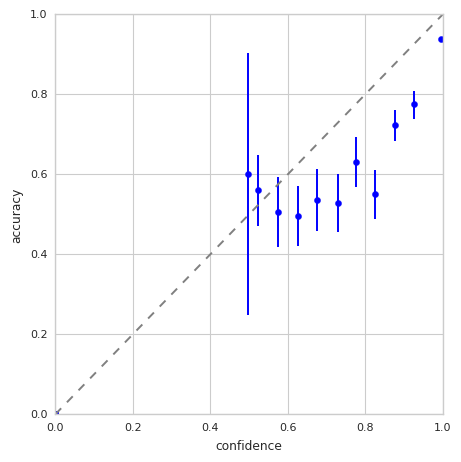} \\

  \vspace{0.2cm}

  \raisebox{0.12\linewidth}{\textbf{c}}
  \includegraphics[width=0.25\linewidth]{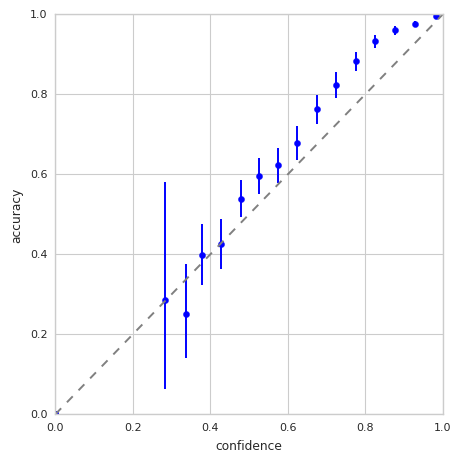}
  \includegraphics[width=0.25\linewidth]{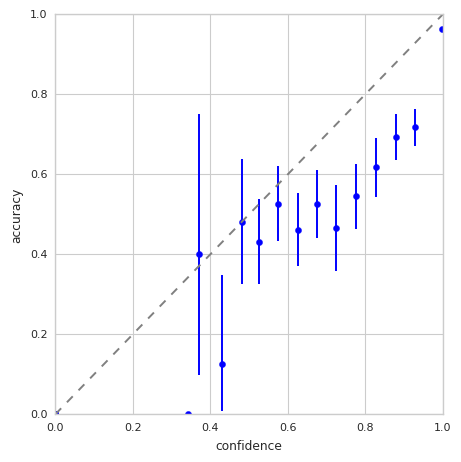}
  \includegraphics[width=0.25\linewidth]{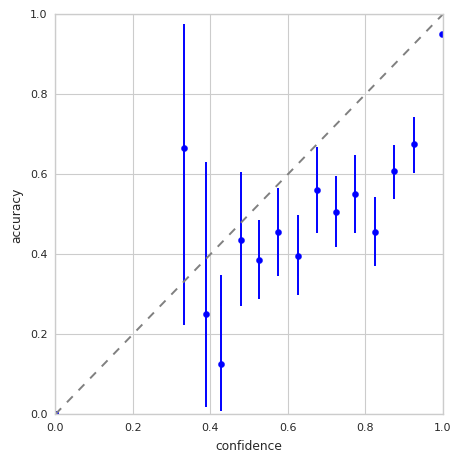} \\
  \raisebox{0.12\linewidth}{\textbf{d}}
  \includegraphics[width=0.25\linewidth]{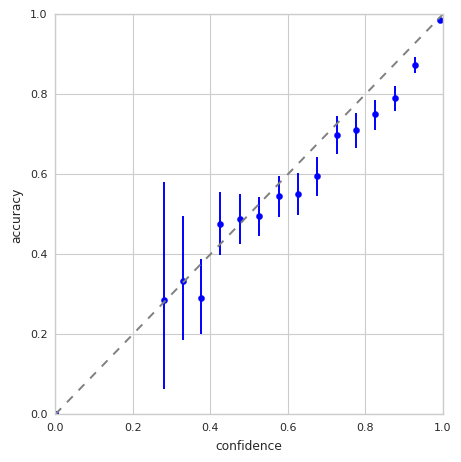}
  \includegraphics[width=0.25\linewidth]{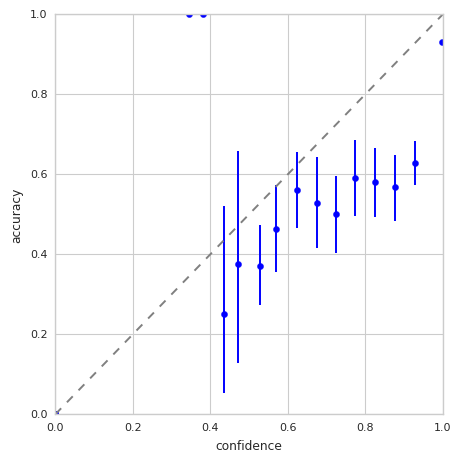}
  \includegraphics[width=0.25\linewidth]{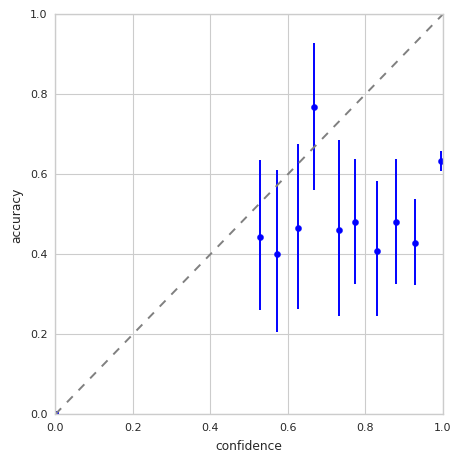} \\

  \caption{
    Calibration plots with 90\% confidence intervals for four of the models after 2,000 steps, 20,000 steps, and 40,000 steps (left, center, and right of each trio):
    \textbf{a} is CEB$_0$;
    \textbf{b} is VIB$_0$;
    \textbf{c} is VIB$_2$;
    \textbf{d} is Determ.
    \textit{Perfect calibration} corresponds to the dashed diagonal lines.
    \textit{Underconfidence} occurs when the points are above the diagonal.
    \textit{Overconfidence} is below the diagonal.
    The $\rho=0$ models are nearly perfectly calibrated still at 20,000 steps, but even at $\rho=2$, the VIB model is almost as overconfident as Determ.
  }
  \label{fig:calibration}
\end{figure}

\paragraph{(RG3): Out-of-distribution detection.}
\label{sec:ood}
We compare the ability of Determ, CEB$_0$, VIB$_0$, and VIB$_4$ to detect four different out-of-distribution (OoD) detection datasets.
$U(0,1)$ is uniform noise in the image domain.
MNIST uses the MNIST test set.
Vertical Flip is the most challenging, using vertically flipped Fashion MNIST test images, as originally proposed in~\citet{uncertainvib}.
CW is the Carlini-Wagner L$_2$ attack~\citep{carliniwagner} at the default settings found in~\citet{cleverhans}, and additionally includes the adversarial attack success rate against each model.

We use two different metrics for thresholding, proposed in~\citet{uncertainvib}.
$H$ is the classifier entropy.
$R$ is the rate, defined in~\Cref{sec:vib}.
These two threshold scores are used with the standard suite of proper scoring rules~\citep{lee2018simple}: \textit{False Positive Rate at 95\% True Positive Rate} (FPR \@ 95\% TPR), \textit{Area Under the ROC Curve} (AUROC), and \textit{Area Under the Precision-Recall Curve} (AUPR).

\Cref{tab:ood} shows that using $R$ to detect OoD examples can be much more effective than using classifier-based approaches.
The deterministic baseline model is far weaker at detection using $H$ than either of the high-performing stochastic models (CEB$_0$ and VIB$_4$).
Those models both saturate detection performance, providing reliable signals for all four OoD datasets.
However, as VIB$_0$ demonstrates, simply having $R$ available as a signal does not guarantee good detection.
As we saw above, the VIB$_0$ models had noticeably worse classification performance, indicating that they had not achieved the MNI point: $I(Y;Z) < I(X;Z)$ for those models.
These results indicate that for detection, violating the MNI criterion by having $I(X;Z) > I(X;Y)$ may not be harmful, but violating the criterion in the opposite direction \emph{is} harmful.

\paragraph{(RG3): Calibration.}
\label{sec:calibration}
A \textit{well-calibrated} model is correct half of the time it gives a confidence of 50\% for its prediction.
In~\Cref{fig:calibration}, we show calibration plots at various points during training for four models.
Calibration curves help analyze whether models are underconfident or overconfident.
Each point in the plots corresponds to a 5\% confidence bin.
Accuracy is averaged for each bin.
All four networks move from under- to overconfidence during training.
However, CEB$_0$ and VIB$_0$ end up only slightly overconfident, while $\rho=2$ is already sufficient to make VIB and CEB (not shown) nearly as overconfident as the deterministic model.

\paragraph{(RG1): Overfitting Experiments.}
\label{sec:generalization}

We replicate the basic experiment from~\citet{rethinkinggeneralization} by using the images from Fashion MNIST, but replacing the training labels with fixed random labels.
This dataset is \textit{information-free} because $I(X;Y) = 0$.
We use that dataset to train multiple deterministic models, as well as CEB and VIB models at $\rho$ from 0 through 7.
We find that the CEB and VIB models with $\rho < 6$ \textit{never} learn, even after 100 epochs of training, but the deterministic models \textit{always} learn.
After about 40 epochs of training they begin to memorize the random labels, indicating severe overfitting and a perfect \textit{failure} to generalize.

\begin{figure}[tb]
  \center
  \vspace{-0.0cm}
  \includegraphics[width=1.0\linewidth]{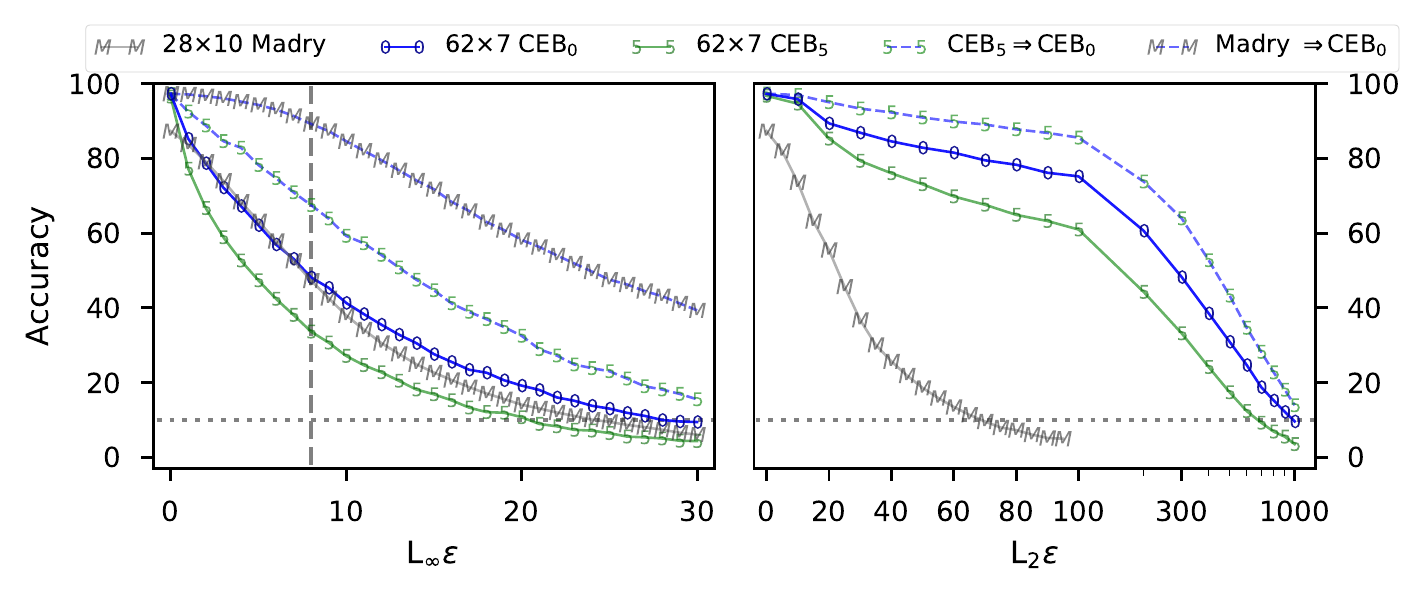}
  \vspace{-0.8cm}
  \caption{%
    \textbf{Left:} Accuracy on untargeted $L_\infty$ attacks at different values of $\varepsilon$ for all 10,000 CIFAR10 test set examples.
    CEB$_0$ is the model with the highest accuracy (97.51\%) trained at $\rho=0$.
    CEB$_5$ is the model with the highest accuracy (97.06\%) trained at $\rho=5$.
    Madry is the best adversarially-trained model from~\citet{madry2017towards} with 87.3\% accuracy (values provided by Aleksander Madry).
    CEB$_5\Rightarrow$CEB$_0$ is transfer attacks from the CEB$_5$ model to the CEB$_0$ model.
    Madry $\Rightarrow$CEB$_0$ is transfer attacks from the Madry model to the CEB$_0$ model.
    Madry was trained with 7 steps of PGD at $\varepsilon=8$ (grey dashed line).
    Chance is 10\% (grey dotted line).
    \textbf{Right:} Accuracy on untargeted $L_2$ attacks at different values of $\varepsilon$.
    All values are collected at 7 steps of PGD.
    CEB$_0$ outperforms Madry everywhere except the region of $L_\infty \varepsilon \in [2,7]$.
    Madry appears to have overfit to L$_\infty$, given its poor performance on $L_2$ attacks relative to either CEB model. 
    \emph{None of the CEB models are adversarially trained.}
  }
  \label{fig:untargeted}
\end{figure}

\subsection{(RG1) and (RG2): CIFAR10 Experiments}
\label{sec:cifar}

For CIFAR10~\citep{cifar} we trained the largest Wide ResNet~\citep{wrn} we could fit on a single GPU with a batch size of 250.
This was a $62 \!\times\! 7$ model trained using AutoAugment~\citep{autoaugment}.
We trained 3 CatGen CEB$_{\text{bidir}}$ models each of CEB$_0$ and CEB$_5$ and then selected the two models with the highest test accuracy for the adversarial robustness experiments.
We evaluated the CatGen models using the consistent classifier, since CatGen models only train $e(z|x)$ and $b(z|y)$.
CEB$_0$ reached \textbf{97.51\%} accuracy.
This result is better than the $28 \!\times\! 10$ Wide ResNet from AutoAugment\ by 0.19 percentage points, although it is still worse than the Shake-Drop model from that paper.
We additionally tested the model on the CIFAR-10.1 test set~\citep{cifar10_1}, getting accuracy of 93.6\%.
This is a gap of only \textbf{3.9} percentage points, which is better than all of the results reported in that paper, and substantially better than the Wide ResNet results (but still inferior to the Shake-Drop AutoAugment\ results).
The CEB$_5$ model reached 97.06\% accuracy on the normal test set and 91.9\% on the CIFAR-10.1 test set, showing that increased $\rho$ gave substantially worse generalization.

To test robustness of these models, we swept $\epsilon$ for both PGD attacks (\Cref{fig:untargeted}).
The CEB$_0$ model not only has substantially higher accuracy than the adversarially-trained Wide ResNet from~\citet{madry2017towards} (\textit{Madry}), it also beats the Madry model on both the L$_2$ and the L$_\infty$ attacks at almost all values of $\epsilon$.
We also show that this model is even more robust to two transfer attacks, where we used the CEB$_5$ model and the Madry model to generate PGD attacks, and then test them on the CEB$_0$ model.
This result indicates that these models are not doing ``gradient masking'', a failure mode of some attempts at adversarial defense~\citep{athalye2018obfuscated}, since these are black-box attacks that do not rely on taking gradients through the target model.

\section{Conclusion}

We have presented the Conditional Entropy Bottleneck (CEB), motivated by the Minimum Necessary Information (MNI) criterion and the hypothesis that failures of \textit{robust generalization} are due in part to learning models that retain \emph{too much} information about the training data.
We have shown empirically that simply by switching to CEB, models may substantially improve their robust generalization, including \textbf{(RG1)} higher accuracy, \textbf{(RG2)} better adversarial robustness, and \textbf{(RG3)} stronger OoD detection.
We believe that the MNI criterion and CEB offer a promising path forward for many tasks in machine learning by permitting fast amortized inference in an easy-to-implement framework that improves robust generalization.

\clearpage
\appendix
\newpage
\appendix

Here we collect a number of results that are not critical to the core of the paper, but may be of interest to particular audiences.

\section{Model Details}
\label{sec:model_details}

\subsection{Fashion MNIST}
All of the models in our Fashion MNIST experiments have the same core architecture:
A $7 \times 2$ Wide Resnet~\citep{wideresnet} for the encoder, with a final layer of $D=4$ dimensions for the latent representation, followed by a two layer MLP classifier using ELU~\citep{elu} activations with a final categorical distribution over the 10 classes.

The stochastic models parameterize the mean and variance of a $D=4$ fully covariate multivariate Normal distribution with the output of the encoder.
Samples from that distribution are passed into the classifier MLP.
Apart from that difference, the stochastic models don't differ from Determ during evaluation.
None of the five models uses any form of regularization (e.g., $L_1$, $L_2$, DropOut~\citep{dropout}, BatchNorm~\citep{batchnorm}).

The VIB models have an additional learned marginal, $m(z)$, which is a mixture of 240 $D=4$ fully covariate multivariate Normal distributions.
The CEB model instead has the backward encoder, $b(z|y)$ which is a $D=4$ fully covariate multivariate Normal distribution parameterized by a 1 layer MLP mapping the label, $Y=y$, to the mean and variance.
In order to simplify comparisons, for CEB we additionally train a marginal $m(z)$ identical in form to that used by the VIB models.
However, for CEB, $m(z)$ is trained using a separate optimizer so that it doesn't impact training of the CEB objective in any way.
Having $m(z)$ for both CEB and VIB allows us to compare the rate, $R$, of each model except Determ.

\subsection{CIFAR-10}
For the $62 \!\times\! 7$ CEB CIFAR-10 models, we used the AutoAugment policies for CIFAR-10.
We trained the models for 800 epochs, lowering the learning rate by a factor of 10 at 400 and 600 epochs.
We trained all of the models using Adam~\citep{adam} at a base learning rate of $10^{-3}$.

\subsection{Distributional Families}

Any distributional family may be used for the encoder.
Reparameterizable distributions~\citep{vae,figurnov2018implicit} are convenient, but it is also possible to use the score function trick~\citep{reinforce} to get a high-variance estimate of the gradient for distributions that have no explicit or implicit reparameterization.
In general, a good choice for $b(z|y)$ is the same distributional family as $e(z|x)$, or a mixture thereof.
These are modeling choices that need to be made by the practitioner, as they depend on the dataset.
In this work, we chose normal distributions because they are easy to work with and will be the common choice for many problems, particularly when parameterized with neural networks, but that choice is incidental rather than fundamental.

\section{Mutual Information Optimization}

As an objective function, CEB is independent of the methods used to optimize it.
Here we focus on variational objectives because they are simple, tractable, and well-understood, but any approach to optimize mutual information terms can work, so long as they respect the side of the bounds required by the objective.
For example, both~\citet{oord2018representation,hjelm2018learning} could be used to maximize $I(Y;Z)$.

\subsection{Finitness of the Mutual Information}

The conditions for infinite mutual information given in~\citet{amjad2018learning} do not apply to either CEB or VIB, as they both use stochastic encoders $e(z|x)$.
In our experiments using continuous representations, we did not encounter mutual information terms that diverged to infinity, although it is possible to make modeling and data choices that make it more likely that there will be numerical instabilities.
This is not a flaw specific to CEB or VIB, however, and we found numerical instability to be almost non-existent across a wide variety of modeling and architectural choices for both variational objectives.

\section{Additional CEB Objectives}
\label{sec:bonus_ceb}

Here we describe a few more variants of the CEB objective.

\subsection{Hierarchical CEB}
\label{sec:cebhier}

\begin{figure}
  \center
  \begin{tikzpicture}[scale=1.5]
    \begin{scope}
      \clip (-0.5, 0.0) circle (0.7cm);
      \clip ( 0.5, 0.0) circle (0.7cm);
      \fill[lightgray] ( 0.5, 0.0) circle (0.7cm);
    \end{scope}
    \begin{scope}
      \clip (-0.5, 0.0) circle (0.7cm);
      \clip ( 0.5, 0.0) circle (0.7cm);
      \clip (-0.7,-0.4) circle (0.7cm);
      \fill[gray] (-0.7,-0.4) circle (0.7cm);
    \end{scope}
    \begin{scope}
      \clip (-0.5, 0.0) circle (0.7cm);
      \clip ( 0.5, 0.0) circle (0.7cm);
      \clip (-0.7,-0.4) circle (0.7cm);
      \clip (-0.7,-0.6) circle (0.7cm);
      \fill[darkgray] (-0.7,-0.6) circle (0.7cm);
    \end{scope}
    \draw (-0.5, 0.0) circle (0.7cm) +(0.0, 0.85) node{$H(X)$};
    \draw ( 0.5, 0.0) circle (0.7cm) +(0.0, 0.85) node{$H(Y)$};
    \draw (-0.7,-0.4) circle (0.7cm) +(0.1, 0.85) node{$H(Z_1)$};
    \draw (-0.7,-0.6) circle (0.7cm) +(0.0,-0.85) node{$H(Z_2)$};
  \end{tikzpicture}
  \caption{%
    Information diagram for the basic hierarchical CEB model, $Z_2 \leftarrow Z_1 \leftarrow X \leftrightarrow Y$.
  }
  \label{fig:hiervenn}
\end{figure}
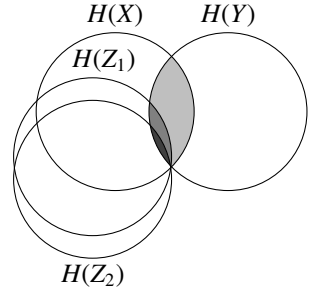

Thus far, we have focused on learning a single latent representation (possibly composed of multiple latent variables at the same level).
Here, we consider one way to learn a hierarchical model with CEB.

Consider the graphical model $Z_2 \leftarrow Z_1 \leftarrow X \leftrightarrow Y$.
This is the simplest hierarchical supervised representation learning model.
The general form of its information diagram is given in~\Cref{fig:hiervenn}.

The key observation for generalizing CEB to hierarchical models is that the target mutual information doesn't change.
By this, we mean that all of the $Z_i$ in the hierarchy should cover $I(X;Y)$ at convergence, which means maximizing $I(Y;Z_i)$.
It is reasonable to ask why we would want to train such a model, given that the final set of representations are presumably all effectively identical in terms of information content.
Doing so allows us to train deep models in a principled manner such that all layers of the network are consistent with each other and with the data.
We need to be more careful when considering the residual information terms, though -- it is not the case that we want to minimize $I(X;Z_i|Y)$, which is not consistent with the graphical model.
Instead, we want to minimize $I(Z_{i-1};Z_i|Y)$, defining $Z_0 = X$.

This gives the following simple \textit{Hierarchical CEB} objective:
\begin{align}
CEB_{\text{hier}} \equiv &\min \sum_i I(Z_{i-1};Z_i|Y) - I(Y;Z_i) \\
\Leftrightarrow &\min \sum_i {-H}(Z_i|Z_{i-1}) + H(Z_i|Y) + H(Y|Z_i)
\end{align}
Because all of the $Z_i$ are targetting $Y$, this objective is as stable as regular CEB.

\subsection{Sequence Learning}
\label{sec:seq}

Many of the richest problems in machine learning vary over time.
In~\citet{predictiveinfo}, the authors define the \textit{Predictive Information}:
$$
I(X_{past}, X_{future}) = \left\langle \log \frac{p(x_{past}, x_{future})}{p(x_{past}) p(x_{future})} \right\rangle
$$
This is of course just the mutual information between the past and the future.
However, under an assumption of temporal invariance (any time of fixed length is expected to have the same entropy),
they are able to characterize the predictive information, and show that it is a subextensive quantity:
$\lim_{T \rightarrow \infty} I(T)/T \rightarrow 0$, where $I(T)$ is the predictive information over a time window of length $2T$ ($T$ steps of the past predicting $T$ steps into the future).
This concise statement tells us that past observations contain vanishingly small information about the future as the time window increases.

The application of CEB to extracting the predictive information is straightforward.
Given the Markov chain $X_{<t} \rightarrow X_{\ge t}$, we learn a representation $Z_t$ that optimally covers $I(X_{<t}, X_{\ge t})$ in \textit{Predictive CEB}:
\begin{align}
CEB_{\text{pred}} \equiv& \min I(X_{<t};Z_t | X_{\ge t}) - I(X_{\ge t}, Z_t) \\
\Rightarrow& \min {-H}(Z_t | X_{<t}) + H(Z_t | X_{\ge t}) + H(X_{\ge t} | Z_t)
\end{align}

Given a dataset of sequences, CEB$_{\text{pred}}$ may be extended to a bidirectional model.
In this case, two representations are learned, $Z_{<t}$ and $Z_{\ge t}$.
Both representations are for timestep $t$, the first representing the observations before $t$, and the second representing the observations from $t$ onwards.
As in the normal bidirectional model, using the same encoder and backwards encoder for both parts of the bidirectional CEB objective ties the two representations together.

\paragraph{Modeling and architectural choices.}
As with all of the variants of CEB, whatever entropy remains in the data after capturing the entropy of the mutual information in the representation must be modeled by the decoder.
In this case, a natural modeling choice would be a probalistic RNN with powerful decoders per time-step to be predicted.
However, it is worth noting that such a decoder would need to sample at each future step to decode the subsequent step.
An alternative, if the prediction horizon is short or the predicted data are small, is to decode the entire sequence from $Z_t$ in a single,
feed-forward network (possibly as a single autoregression over all outputs in some natural sequence).
Given the subextensivity of the predictive information, that may be a reasonable choice in stochastic environments, as the useful prediction window may be small.

Likely a better alternative, however, is to use the CatGen decoder, as no generation of the long future sequences is required in that case.

\paragraph{Multi-scale sequence learning.}
As in WaveNet~\citep{wavenet}, it is natural to consider sequence learning at multiple different temporal scales.
Combining an architecture like time-dilated WaveNet with CEB is as simple as combining CEB$_{\text{pred}}$ with CEB$_{\text{hier}}$~(\Cref{sec:cebhier}).
In this case, each of the $Z_i$ would represent a wider time dilation conditioned on the aggregate $Z_{i-1}$.

\subsection{Unsupervised CEB}

For unsupervised learning, it seems challenging to put the decision about what information should be kept into objective function hyperparameters,
as in the $\beta$ VAE and penalty VAE~\citep{brokenelbo} objectives.
That work showed that it is possible to constrain the amount of information in the learned representation, but it is unclear
how those objective functions keep only the ``correct'' bits of information for the downstream tasks you might care about.
This is in contrast to supervised learning while targeting the MNI point, where the task clearly defines the both the correct amount of information and which bits are likely to be important.

Our perspective on the importance of defining a task in order to constrain the information in the representation suggests that we can turn the problem
into a data modeling problem in which the practitioner who selects the dataset also ``models'' the likely form of the useful bits in the dataset for the
downstream task of interest.

In particular, given a dataset $X$, we propose selecting a function $f(X) \rightarrow X'$ that transforms $X$ into a new random variable $X'$.
This defines a paired dataset, $P(X,X')$, on which we can use CEB as normal.
Note that choosing the identity function for $f$ results in maximal mutual information between $X$ and $X'$ ($H(X)$ nats), which will result in a representation that is far from the MNI for normal downstream tasks.

It may seem that we have not proposed anything useful, as the selection of $f(.)$ is unconstrained, and seems much more daunting than selecting $\beta$ in a $\beta$ VAE or $\sigma$ in a penalty VAE.
However, there is a very powerful class of functions that makes this problem much simpler, and that also make it clear using CEB will \textit{only} select bits from $X$ that are useful.
That class of functions is the noise functions.

\subsubsection{Denoising CEB Autoencoder}

\begin{figure}
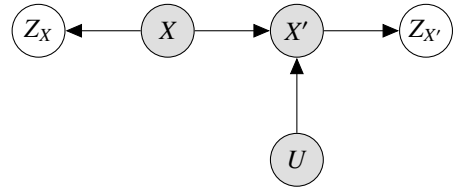

  \center
  \tikz{ %
    \node[latent] (ZX) {$Z_X$} ; %
    \node[obs, right=of ZX] (X) {$X$} ; %
    \node[obs, right=of X] (XP) {$X'$} ; %
    \node[obs, below=of XP] (U) {$U$} ; %
    \node[latent, right=of XP] (ZXP) {$Z_{X'}$} ; %
    \edge {X} {ZX} ; %
    \edge {XP} {ZXP} ; %
    \edge {U} {XP} ; %
    \edge {X} {XP} ; %
  }

  \caption{%
    Graphical model for the Denoising CEB Autoencoder.
  }
  \label{fig:unsupervisedgraph}
\end{figure}

Given a dataset $X$ without labels or other targets, and some set of tasks in mind to be solved by a learned representation, we may select a random noise variable $U$, and function $X' = f(X,U)$
that we believe will destroy the irrelevant information in $X$.
We may then add representation variables $Z_X, Z_{X'}$ to the model, giving the joint distribution $p(x,x',u,z_X,z_{X'}) \equiv p(x)p(u)p(x'|f(x,u))e(z_X|x)b(z_{X'}|x')$.
This joint distribution is represented in~\Cref{fig:unsupervisedgraph}.

\textit{Denoising Autoencoders} were originally proposed in~\citet{vincent2008extracting}.
In that work, the authors argue informally that reconstruction of corrupted inputs is a desirable property of learned representations.
In this paper's notation, we could describe their proposed objective as $\min H(X|Z_{X'})$, or equivalently $\min \left\langle \log d(x|z_{X'}=f(x,\eta)) \right\rangle_{x,\eta \sim p(x)p(\theta)}$\,.

We also note that, practically speaking, we would like to learn a representation that is consistent with uncorrupted inputs as well.
Consequently, we are going to use a bidirectional model.
\begin{align}
CEB_{\text{denoise}} \equiv& \min I(X;Z_X|X') - I(X';Z_X) \\
&\quad \quad + I(X';Z_{X'}|X) - I(X;Z_{X'}) \\
\Rightarrow& \min {-H}(Z_X|X) + H(Z_X|X') + H(X'|Z_X) \\
&\quad \quad - H(Z_{X'}|X') + H(Z_{X'}|X) + H(X|Z_{X'})
\end{align}
This requires two encoders and two decoders, which may seem expensive, but it permits a consistent learned representation that can be used cleanly for downstream tasks.
Using a single encoder/decoder pair would result in either an encoder that does not work well with uncorrupted inputs, or a decoder that only generates noisy outputs.

If you are only interested in the learned representation and not in generating good reconstructions, the objective simplifies to the first three terms.
In that case, the objective is properly called a \textit{Noising CEB Autoencoder}, as the model predicts the noisy $X'$ from $X$:
\begin{align}
CEB_{\text{noise}} \equiv& \min I(X;Z_X|X') - I(X';Z_X) \\
\Rightarrow& \min -H(Z_X|X) + H(Z_X|X') + H(X'|Z_X)
\end{align}

In these models, the noise function, $X' = f(X,U)$ must encode the practitioner's assumptions about the structure of information in the data.
This obviously will vary per type of data, and even per desired downstream task.

However, we don't need to work too hard to find the perfect noise function initially.
A reasonable choice for $f$ is:
\begin{align}
f(x,\eta) =& \, \text{clip}(x + \eta, \mathcal{D}) \\
\eta \sim& \, \lambda U(-1,1) * \mathcal{D} \\
\mathcal{D} =& \, \text{domain}(X)
\end{align}
In other words, add uniform noise scaled to the domain of $X$ and by a hyperparameter $\lambda$, and clip the result to the domain of $X$.
When $\lambda = 1$, $X'$ is indistinguishable from uniform noise.
As $\lambda \rightarrow 0$, this maintains more and more of the original information from $X$ in $X'$.
For some value of $\lambda > 0$, most of the irrelevant information is destroyed and most of the relevant information is maintained,
if we assume that higher frequency content in the domain of $X$ is less likely to contain the desired information.
That information is what will be retained in the learned representation.

\paragraph{Theoretical optimality of noise functions.}
Above we claimed that this learning procedure will only select bits that are useful for the downstream task, given that we select the proper noise function.
Here we prove that claim constructively.
Imagine an oracle that knows which bits of information should be destroyed, and which retained in order to solve the future task of interest.
Further imagine, for simplicity, that the task of interest is classification.
What noise function must that oracle implement in order to ensure that $CEB_{denoise}$ can only learn exactly the bits needed for classification?
The answer is simple: for every $X=x_i$, select $X'=x_i'$ uniformly at random from among all of the $X=x_j$ that should have the same class label as $X=x_i$.
Now, the only way for CEB to maximize $I(X;Z_{X'})$ and minimize $I(X';Z_{X'})$ is by learning a representation that is isomorphic to classification, and that encodes exactly $I(X;Y)$ nats of information, even though it was only trained ``unsupervisedly'' on $X,X'$ pairs.
Thus, if we can choose the correct noise function that destroys only the bits we don't care about, $CEB_{denoise}$ will learn the desired representation and nothing else (caveated by model, architecture, and optimizer selection, as usual).

\bibliography{bib}
\bibliographystyle{icml2020}

\end{document}